\pdfoutput=1

\documentclass[11pt]{article}

\usepackage[]{ACL2023}

\usepackage{times}
\usepackage{latexsym}

\usepackage[T1]{fontenc}

\usepackage[utf8]{inputenc}

\usepackage{microtype}

\usepackage{inconsolata}
\usepackage{graphicx}
\usepackage{placeins}

\usepackage{hyperref}
\usepackage{float}

\title{Gender-Inclusive Grammatical Error Correction through Augmentation}

\author{Gunnar Lund, Kostiantyn Omelianchuk, Igor Samokhin \\ Grammarly \\ \texttt{\{gunnar.lund,kostiantyn.omelianchuk,igor.samokhin\}@grammarly.com}}

\begin{document}
\maketitle
\begin{abstract}
In this paper we show that GEC systems display gender bias related to the use of masculine and feminine terms and the gender-neutral singular \textit{they}. We develop parallel datasets of texts with masculine and feminine terms, and singular \textit{they}, and use them to quantify gender bias in three competitive GEC systems. We contribute a novel data augmentation technique for singular \textit{they} leveraging linguistic insights about its distribution relative to plural \textit{they}. We demonstrate that both this data augmentation technique and a refinement of a similar augmentation technique for masculine and feminine terms can generate training data that reduces bias in GEC systems, especially with respect to singular \textit{they} while maintaining the same level of quality.
\end{abstract}

\section{Introduction}

Natural Language Processing (NLP) systems are well known to exhibit sensitivity to social characteristics, a sensitivity that may lead to harms for users interacting with these systems. In this work, we examine how NLP systems performing the task of Grammatical Error Correction (GEC) are sensitive to \textit{gender} characteristics in English and how this sensitivity represents bias that harms users of these systems. We propose a new data augmentation technique to address bias related to singular \textit{they}, and show how it can mitigate gender bias. We also show how this technique interacts with prior work on data augmentation methods to reduce bias.

The expression of gender in English is complex, and to focus our study, we identify discrete biases that GEC systems may exhibit. First, we build on other works on gender bias in NLP by examining how masculine and feminine terms impact the behavior of GEC systems. Discrepant behavior between sentences that contain one or the other is evidence of bias. Second, we examine the behavior of GEC systems with a gender-neutral pronominal paradigm in English commonly called singular \textit{they}. The linguistic properties of this paradigm introduce additional kinds of biases relative to masculine or feminine pronouns. Additionally, we make our evaluation datasets publicly available\footnote{\url{https://github.com/grammarly/gender-inclusive-gec}}.

To prevent these adverse behaviors, we apply techniques to generate synthetic training data that address different aspects of these behaviors. First, we adopt Counterfactual Data Augmentation (CDA), which is used successfully to reduce bias in word embedding models \citep{lu_gender_2019, maudslay_its_2020}, and apply it to the GEC case. Second, we introduce a new technique for generating training data with \textit{they} pronouns that have unambiguously singular reference using insight from theoretical linguistics to target model bias with gender-neutral sentences. Because these techniques are data-oriented, the innovations should theoretically generalize beyond the GEC domain to other NLP tasks as well.

Our main contributions are:
\begin{itemize}
    \item We introduce a novel technique for creating singular \textit{they} data, leveraging specific linguistic features of this use of the pronoun. Additionally, we refine previous approaches to addressing discrepancies in model behavior between texts with masculine and feminine pronouns and show their application outside of the context of masked language models.
    \item We qualitatively and quantitatively measure biases in competitive GEC systems by comparing model performance on parallel test sets containing singular \textit{they} pronouns, masculine terms, and feminine terms.
    \item We show how our data augmentation techniques, both in isolation and combination, mitigate these biases when used to create training data for these GEC systems with a minimal impact to overall performance.
\end{itemize}

\section{Background}

\subsection{Gender and Bias in NLP and GEC}
\subsubsection{Conceptual grounding}
To orient this work, we highlight two recent calls-to-action regarding the study of bias and gender in NLP systems. First, following \citet{devinney_theories_2022}, we mean to be explicit about the conception of gender and gendered language we assume. In particular, we are concerned with gendered linguistic \textit{content}, and not the gender of the authors or readers of that content. We recognize that the expression of gender in English is notional; the use of some nouns and pronouns is linked to particular gendered conceptual categories \citep{mcconnell-ginet__2013, ackerman_syntactic_2019}. Additionally, the \textit{use} of language with gendered content represents one aspect of gender performativity which produces and reifies these gendered categories. 

Second, following \citet{blodgett_language_2020}, “bias” is an inherently normative concept. In the context of NLP systems, it must be understood in terms of the potential \textit{harms} that those systems may cause and to \textit{whom} those harms may be caused. Therefore, we directly focus on mitigating harms themselves as they relate to the GEC task and how users of these systems interact with them and may be affected by them. Unlike some studies of bias in upstream contexts like word embeddings, users interact with GEC systems directly; these users choose to incorporate these systems’ suggestions into their emails, essays, and tweets, and bias may impact anyone interacting with such text.

Further, because the GEC task is an inherently normative one on its own—these systems offer suggestions to \textit{correct} a user’s text and are designed in accordance with preexisting normative notions of “correct” or “fluent” English—GEC systems necessarily also participate in the production of gendered categories. The norms assumed when constructing these systems and datasets in this regard may conflict with other norms about language use. For example, there are norms against the use of singular \textit{they} in some language communities. Some English speakers do not accept singular \textit{they} as a grammatical construction of English \citep[a.o.]{bjorkman_singular_2017}, and some prescriptive grammars advise against the use of singular \textit{they} (c.f., \citealp{strunk_elements_1999}). People who are non-binary and use \textit{they} pronouns, e.g., cannot refer to themselves "correctly" within these circumscribed norms of language use. As we discuss below, the operationalization of these norms in GEC systems may lead to harm. We adopt the view that GEC systems should reflect the most permissive distribution of singular \textit{they}. This distribution is discussed further in section \ref{sec:st-cda}.

Our work has a notable limitation in that we do not investigate bias with respect to neopronouns like  \textit{ze} or \textit{xe}. We leave extensions of CDA-like techniques to these pronouns for future work.

\subsubsection{Two biases}
We identify two areas where GEC systems can produce biased, and therefore potentially harmful, outcomes. Importantly, this is not an exhaustive account of potential biases GEC systems exhibit, but we think this is as good a starting point as any.

First, a GEC system can be implicitly biased if it consistently performs better on texts containing words of one gendered category over another. This is an allocative harm. If a GEC system performs worse on texts that are about people who use one pronoun or another, texts about those people may contain more grammatical errors, impacting their relative opportunity. For example, a user writing letters of recommendation may inadvertently include more grammatical errors in letters for individuals using masculine pronouns, as a system could perform worse on texts with masculine pronouns than feminine pronouns, and this could impact the relative reception those letters receive compared to similar letters with feminine pronouns.

Second, a GEC system can be explicitly biased if it offers corrections that reify harmful notions about particular gendered categories, including the reinforcement of stereotypes and misgendering or erasure of individuals referred to in the user's text. This is a representational harm. This harm is explicitly called out by participants in a survey on harms of AI systems with respect to non-binary individuals \citep{dev_harms_2021}. The examples below are representative of these kinds of corrections. The first is an instance of misgendering, replacing singular \textit{they} with a masculine pronoun; the second is an instance of erasure, implying that \textit{they} has a correct use only as a plural pronoun.

\begin{enumerate}
    \item I asked Alex \textbf{their} phone number. -> I asked Alex \textbf{his} phone number.
    \item They are \textbf{a linguist}. -> They are \textbf{linguists}.
\end{enumerate}

We find evidence for both of these biases by analyzing the following GEC systems (table \ref{short-table-sota-overview}) 
\begin{enumerate}
\item  GECToR \cite{omelianchuk_gector_2020}, sequence tagging approach, which was a  state-of-the-art GEC model in 2020
\item Fine-tuned BART model \cite{lewis-etal-2020-bart},  which represents another popular and competitive approach - sequence to sequence
\item  EditScorer \cite{sorokin-2022-improved}, the recent ranker approach, that is the second-best result on BEA Shared Task 2019, as of April 2023\footnote{\url{http://nlpprogress.com/english/grammatical_error_correction.html}}.
\end{enumerate}

Our quantitative analysis revealed that, for all three systems, there is a significant gap (from -6.2\% to -9.5\% F05 points) between the original and augmented with singular \textit{they} examples versions of the BEA-dev subset, which we call bea-195.
The details on how we built these datasets and evaluation approach is provided in sections \ref{sec:datasets} and \ref{sec:evaluation}. Detailed evaluation results are available in appendix table \ref{full-table-sota-overview}. 

\begin{table*}[h]
\centering
\resizebox{\textwidth}{!}{
\begin{tabular}{|c|c|ccc|ccc|}
\hline
\textbf{System} & 
\multicolumn{1}{c|}{\textbf{bea-dev-full}} & 
\multicolumn{3}{c|}{\textbf{bea-195}} & 
\multicolumn{3}{c|}{\textbf{bea-556}} 
\\ 
\cline{2-8}
 &  
 \textbf{F05}  & 
 \textbf{F05 orig}   & \textbf{F05 st aug} & \textbf{diff} & 

  \textbf{F05 orig}   & \textbf{F05 mf aug} & \textbf{diff}  \\ \hline

GECToR (roberta-base) & 54.57\% & 58.28\%  & 48.74\% & -9.54\%   & 59.23\%  & 58.96\% & -0.27\% \\

BART (seq2seq)  &  52.74\%   & 56.36\%  & 50.13\%  & -6.23\%   &  58.61\% & 58.79\% & 0.18\% \\

EditScorer (roberta-large)  & 58.92\% & 60.6\% & 54.16\% & -6.44\%  & 62.55\%  & 61.59\% & -0.96\%\\

\hline

\end{tabular}
}
\caption{\label{short-table-sota-overview} Scores on BEA-dev subsets for strong GEC baselines.}
\end{table*}

We hypothesize that these biases share a partial cause: an imbalance in the training data. If, e.g., the training data with masculine words is of a higher quality than that with feminine words, there may be a performance gap. In the case of unnecessary corrections of singular \textit{they}, we hypothesize that the imbalance is caused by an extreme lack of singular \textit{they} sentences relative to plural \textit{they} sentences. In the remainder of this paper, we show that the introduction of synthetic data helps mitigate these biases.

\subsection{Related work}
\subsubsection{Data augmentation}
Data augmentation has been used in other NLP domains to mitigate gender bias, but most of these works focus on just the masculine and feminine gender categories in English and limit the application of these techniques to word embedding models. \citet{zhao_gender_2018, rudinger_gender_2018, lu_gender_2019} show that coreference resolution systems are sensitive to masculine and feminine words in otherwise equivalent sentences. \citet{lu_gender_2019} use what they call Counterfactual Data Augmentation (CDA) to reduce this sensitivity. In CDA, masculine pronouns are swapped for feminine ones and vice versa. They also swap definitionally gendered common nouns like \textit{actor} and \textit{actress}. They set aside data where the swapping candidates are in a cluster with a proper name.

\citet{maudslay_its_2020} extend \citet{lu_gender_2019} CDA and implement additional name swapping, where the gendered associations of names were determined using census data from the US Social Security Administration. They use this technique to minimize gendered differences in word embedding spaces as measured by WED \citep{bolukbasi_man_2016}. As in \citeauthor{lu_gender_2019}, \citeauthor{maudslay_its_2020} limit their method to masculine and feminine categories.

In addition, gender-neutral data augmentation methods have been proposed in concurrent works by \citet{sun_they_2021} and \citet{vanmassenhove_neutral_2021}. These methods have different goals than ours and are designed to produce different kinds of data. We discuss the differences between these methods and our own in section \ref{sec:differences}.

To our knowledge, ours is the first work to use CDA techniques to reduce bias in GEC systems and the first to use singular \textit{they} augmentation to inject synthetic training data to reduce bias with singular \textit{they} sentences.

\subsubsection{Singular \textit{they} and NLP systems}
Previous works investigating bias towards singular \textit{they} sentences have generally focused on coreference resolution systems. \citet{cao_toward_2021} develop a dataset to evaluate these systems on naturalistic texts about individuals who identify as non-binary, where 35\% of the pronouns are singular \textit{they}. They report that the Stanford system is the highest scoring on this dataset with an F1 score of 34.3\%. This same system reports a much higher F1 score of 60\% on the CONLL 2012 test set.

\citet{baumler_recognition_2022} compare coreference resolution system performance directly on singular \textit{they} sentences compared to plural \textit{they} sentences along the lines of the Winograd or Winogender schemata \citep{levesque_winograd_2012, rudinger_gender_2018, zhao_gender_2018}. They find across-the-board gaps in system performance between the two test sets.

Outside of coreference resolution, \citet{dev_harms_2021} investigate biased representations with BERT in a masked word prediction task. They find that for masked pronouns, BERT has a high accuracy in the prediction of masculine and feminine pronouns, but accuracy considerably lowers for singular \textit{they}.

\section{Description of the data augmentation methods}
We use two data augmentation methods. First, we follow \citet{lu_gender_2019} and others in swapping out feminine words for masculine words and vice versa. Second, we propose a novel augmentation method for generating singular \textit{they} data from sentences containing masculine and feminine pronouns. We treat singular \textit{they} differently because language internal facts about English necessitate a separate treatment: unlike \textit{he} and \textit{she}, \textit{they} has a second life as a plural pronoun.

\subsection{Feminine/Masculine CDA (FM-CDA)}
Consistent with masculine/feminine-term swapping methods in other works, we swap three kinds of nominal terms: 
\begin{itemize}
    \item Pronouns: Swap masculine pronouns for their feminine counterparts and vice versa. Ex: \textit{him} $\rightarrow$ \textit{her}. Because the masculine and feminine pronominal paradigms are partly syncretic—the feminine pronoun \textit{her} can be accusative or possessive and map to \textit{him} or \textit{his}, respectively—token POS tags, generated by a proprietary POS tagger, were used to appropriately match terms to their case-same counterpart.
    \item Common nouns: Swap definitionally feminine common nouns for their masculine gendered counterparts and vice versa. Ex: \textit{actor} $\rightarrow$ \textit{actress}. The selection and mapping of these nouns were hand-curated by industry experts.
    \item Names: Swap first names that are usually associated with feminine terms for names usually associated with masculine terms and vice versa. We partnered with industry experts to curate dictionaries of masculine and feminine names. Because names don’t necessarily have gendered counterparts in the way that pronouns or common nouns do, an arbitrary mapping of names was created. Names occurring in both lists were excluded from swapping.
\end{itemize}

Unlike some previous work involving CDA on fully unsupervised tasks where the creation of a single counterpart sentence is sufficient, GEC training data consists of pairs of ungrammatical source text and grammatical target text. This introduces challenges similar to those that CDA faces for machine translation data consisting of parallel texts \citep{saunders-byrne-2020-reducing, wang_measuring_2022} Because, e.g., the POS tagger may perform differently on the two texts, especially given that the source text is ungrammatical, the swapping algorithm may produce inconsistent swaps if applied separately to the source and target texts. This inconsistency can introduce grammatical errors between the source and target texts and negatively impact model performance. 

To avoid this, an additional algorithm ensures a consistent swap between the source and target text where possible. We first apply our algorithm to the grammatically corrected target text. Then we use an alignment algorithm to align the target and source texts and isolate the differing segments of the texts. For each differing segment, we determine if the number of tokens in the source and target segments is the same, and if not, we discard the data. Then we compare every token in the source and target segments; if the source word would have the same swap as the target, we replace the source word with its differently gendered counterpart. If neither word is swappable, we do nothing. If there is a mismatch in the swap between source word and target word, we discard the data point. For the evaluation set discussed below, we reintroduced this discarded data and manually edited the data to introduce singular \textit{they} pronouns and ensure that the results are parallel.

\subsection{Singular \textit{they} CDA (St-CDA)}\label{sec:st-cda}
Singular \textit{they} is an inherently referential phenomenon. As is evident in the name, it is distinguished from plural \textit{they} because it refers to singular individuals. Further, theoretical and experimental linguistic works show that the overall distribution of singular \textit{they} is conditioned by the nature of the singular antecedent and discourse participants’ relation to the antecedent \citep{bjorkman_singular_2017, ackerman_syntactic_2019, moulton_singular_2020, konnelly_gender_2020, han_processing_2022}. Speakers may be sensitive to linguistic vs. non-linguistic antecedents \citep{moulton_singular_2020}, specificity and definiteness of the antecedent \citep{bjorkman_singular_2017, konnelly_gender_2020}, the discourse participant’s knowledge of the referent’s gender identity \citep{bjorkman_singular_2017, ackerman_syntactic_2019, konnelly_gender_2020}, and the association of a lexical item or name with a particular gender category \citep{bjorkman_singular_2017, ackerman_syntactic_2019, moulton_singular_2020}. In the case of the broadest distribution, singular \textit{they} is used in the same ways that masculine and feminine pronouns are used—the referent’s preference largely dictates the choice of pronoun–but \textit{they} may additionally be used when the referent’s preference is not known \citep{konnelly_gender_2020}.

Differently than feminine and masculine pronouns, we hypothesize that adverse model behavior with singular \textit{they} is at least partially caused by its infrequency relative to plural \textit{they}. Therefore, it is not enough to simply create data that has \textit{they} pronouns; it must also be evidently singular as well. We leverage these linguistic insights about antecedenthood to identify contexts where swapping will result in unambiguous cases of singular \textit{they} by identifying singular antecedents of the pronouns in the text.

We implement this by using HuggingFace’s Neuralcoref coreference resolution system\footnote{\url{https://github.com/huggingface/neuralcoref}} built on top of SpaCy\footnote{Neuralcoref works with SpaCy v2.1 (\url{https://v2.spacy.io/}).}. For a given coreference cluster with a masculine or feminine pronoun, we look at the coreferring expressions in the cluster. If we find a singular one, we perform the swap. We consider a coreferring expression singular if it is:
\begin{enumerate}
    \item A singular common or proper noun.
    \item A singular possessum (e.g., \textit{his foot}).
\end{enumerate}

In addition, \textit{they} has different verbal agreement paradigms than \textit{he} or \textit{she}. We resolve this by using SpaCy’s dependency parser to identify agreeing verbs with the swapped pronouns. We then use the pyInflect package\footnote{\url{https://github.com/bjascob/pyInflect}} to select the verbal inflection consistent with subject agreement with \textit{they}. Finally, we use POS information to disambiguate syncretic forms of \textit{her}. For reflexive pronouns, we swap in the form \textit{themself} and not \textit{themselves} as this form is much less common in the preexisting training data and is more likely to lead to a singular interpretation of \textit{they} pronouns.

As in the case of FM-CDA, we face the potential for inconsistency if this technique is used separately for both source and target data. We use the same algorithm as FM-CDA to perform safe swaps with an additional check on the verbs that were corrected to agree with \textit{they} in the target swapped data.

\begin{figure}
    \centering
    \includegraphics[scale=0.25]{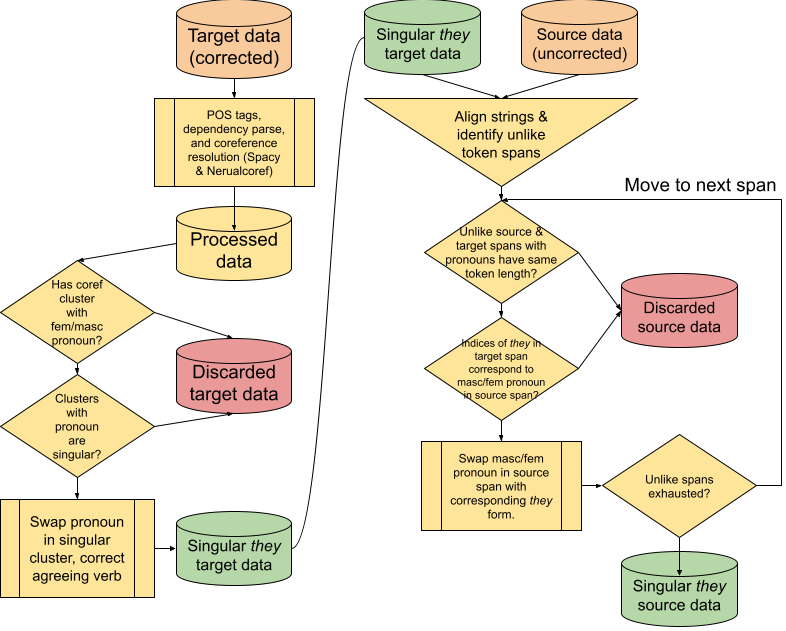}
    \caption{A graphical illustration of the St-CDA swapping process.}
    \label{fig:swap}
\end{figure}

\subsection{Differences between St-CDA and other approaches}\label{sec:differences}
To our knowledge, there are two similar approaches to generating singular \textit{they} data in \cite{sun_they_2021} and \cite{vanmassenhove_neutral_2021}. Their approaches differ from ours in two crucial ways. First, their techniques are meant to create wholly gender-neutral texts, swapping out instances of definitionally gendered noun phrases like “fireman” for “firefighter”. We do not perform these swaps because \textit{they} pronouns may corefer with definitionally gendered words, and this data, in particular, is likely to be rare in most corpora. As we see qualitatively, the baseline model we test seems to be particularly likely to “correct” singular \textit{they} sentences unnecessarily when there is a definitionally gendered word in the sentence.

Second, and most crucially, we hypothesize that the performance gap with singular \textit{they} sentences is due to the relative glut of plural \textit{they} data compared to singular \textit{they} data. As such, we seek to add \textit{they} data that unambiguously has singular reference. Their techniques, by contrast, may result in data where \textit{they} may have a primarily plural interpretation. By targeting contexts where \textit{they} is more likely to be interpreted singularly, we believe St-CDA produces data that will have a higher positive impact and have fewer adverse effects on model quality.

Ultimately, \citet{sun_they_2021} and \citet{vanmassenhove_neutral_2021} have different goals than we do, and this informs the differences in our techniques. Both works envision their technique to be used at runtime in machine translation tasks to, e.g., ensure translations from languages with grammatical gender result in gender neutral translations in English. While they speculate that their techniques can be used to create augmented training data, as we do in this paper, they do not specify what issues they intend this augmented training data to address. By contrast, we seek to counteract a particular imbalance between plural and singular \textit{they} sentences.

\section{Experiments with GEC}

\subsection{Description of datasets}\label{sec:datasets}

For training data, we chose a large Lang-8 Corpus of Learner English \citep{mizumoto_mining_2011}, and more specifically, its “cleaned” version cLang-8 \citep{rothe_simple_2021} which contains over 2 million corrected English sentences. The downside is the noisiness of the data (even in the “cleaned” version) and the lack of consistency in annotations. There are few other GEC datasets of comparable size \citep{bryant_grammatical_2022}. Naturally, not every sentence contains personal pronouns, so only a subset of the dataset is suitable for data augmentation. The size of cLang-8 allowed us to produce about 63 thousand sentences with singular-they augmentation and 254 thousand gender-swapped sentences, which is enough for fine-tuning purposes. In our further experiments, we used only a random sample of 50 thousand sentences from each augmented version of the data to make sure that the results were not impacted by a difference in data size. 

\subsection{Evaluation approach}\label{sec:evaluation}
\subsubsection{Description of the evaluation procedure}

There is no evaluation set which would specifically contain multiple uses of the singular they, so we need to apply data augmentation here as well. To do this, we use the dev part of the BEA-2019 shared task \citep{bryant_bea-2019_2019} since it is one of the standard evaluation sets for GEC. Of 4384 sentences in the BEA dataset, 195 singular \textit{they} sentences were created by replacing the pronouns “he” and “she” with singular “they.” To do this, we applied the CDA-st algorithm described above. The data discarded by the alignment algorithm was also collected and manually revised where possible (sentences where, e.g., a pronoun was inserted or changed from one gendered pronoun to another were either eliminated or revised to eliminate the error). Finally, the entire dataset was manually reviewed to ensure consistency between the original data and the augmented data.

To find the difference in GEC performance on sentences with and without the singular “they,” we evaluate on the subset of 195 sentences before augmentation, “BEA-195-orig”, and on the 195 augmented sentences, “BEA-195-st-aug”. The dataset size limits the conclusions we can make about the GEC model's performance in general, but the differences between scores obtained by the GEC models on these two subsets are statistically significant.

We repeat this procedure for experiments involving masculine and feminine swapping. In this case, our augmentation produced subsets of 556 sentences: “BEA-556-orig” and “BEA-556-mf-aug”.

\subsubsection{Error distribution analysis}
To ensure that our augmentation did not affect edits and shift the error distribution, we conducted a qualitative analysis of m2 files produced by Errant tool on parallel sentences of the original and augmented versions of "195" and "556" evalsets. As shown in the edit type distribution (appendix section \ref{sec:errortypedist}), there are only minor differences in the number of edits (less than 1\% of edits affected). It can be explained by the fact that sometimes Errant might represent similar edits by single or multiple edits, like in the following example \ref{aug_example_table}. \\
The error type distribution for both subsets is available in appendix section \ref{sec:errorcategory}.

\begin{table*}
\centering
\begin{tabular}{|c|c|c|}
\hline
\textbf{Data source} & \textbf{Sentence} & \textbf{Edits}  \\
\hline
original bea-dev &
I love this game because my favourite & sport man belong to this \\
& sport \textbf{man} belong to this game . &  => sportsman plays \\
\hline
mf aug bea-dev & I love this game because my favourite & sport woman  => sportswoman \\
& sport \textbf{woman} belong to this game . &  belong to this => plays \\

\hline
\end{tabular}
\caption{\label{aug_example_table}
An example of a sentence with a different number of edits in an m2 file depending on data augmentation.
}
\end{table*}

\subsubsection{Questions to answer with evaluation}

Running evaluation on these datasets, we are interested mainly in answering two questions:
\begin{itemize}
    \item Is the state-of-the-art GEC model, which was not trained specifically with singular “they” or gender-swapped data, producing worse corrections on the augmented evaluation dataset?
    \item If the corrections are worse, can we shrink or remove the gap in performance by fine-tuning the model on the augmented training data?
\end{itemize}

\subsection{Description of models}
For experiments, we use GECToR \citep{omelianchuk_text_2021} - a state-of-the-art GEC model based on the efficient sequence tagging approach to corrections. Instead of producing a new error-free sentence, GECToR predicts a sequence of tags denoting operations: \emph{“keep,” “remove,” “insert\_X,”} or \emph{“append\_X.”} The corrected text is reconstructed from the original sentence and the tags. Sequence tagging is computationally cheaper than autoregressive approaches, which makes GECToR up to ten times faster than sequence-to-sequence models. At the same time, GECToR set the state-of-the-art at the time of publication.

GECToR is trained and fine-tuned in several stages, starting from the pre-trained language model such as RoBERTa \citep{liu_roberta_2019}. One can also start from the fine-tuned GECToR checkpoint (available on GitHub) and fine-tune it further on the data specifically tailored to the task at hand. However, it may lead the catastrophic forgetting issue, and the overall performance of the model on the general GEC test sets may deteriorate.

\subsection{Experiment approach}
We select GECToR for our fine-tuning experiments due to it being a competitive GEC system and having code that is publicly available. We use weights of the pre-trained GECToR (with RoBERTa-base encoder) model as initialization and fine-tune it for 5 epochs on the following data:

\begin{enumerate}
    \item Original clang8 sentences (~2.2m sentences)
    \item Mix of original and augmented clang8 sentences of one type (~2.2m + 50k sentences, either singular-they or gender-swapped)
    \item Mix of original and augmented clang8 sentences of both types (~2.2m + 100k sentences, 50k for both singular-they and gender-swapped)
\end{enumerate}

We fine-tune the model for 5 epochs with early stopping after 3 epochs and 1 cold epoch. For each training data configuration, we run training 10 times with different random seeds and report the average across all run results. The full list of hyperparameters for fine-tuning can be found in Appendix B. 

Because the baseline GECToR model is already strong enough (it was a SOTA model in 2020) and clang-8 is high-quality data produced by another strong GEC system gT5 xxl \citep{rothe_simple_2021}, the fine-tuning does not lead to substantial quality degradation. As shown in table \ref{table-bea-dev-full}, the differences in F0.5 scores are statistically insignificant.

\begin{table}[h]
\centering
\resizebox{0.5\textwidth}{!}{
\begin{tabular}{|c|ccc|c|}
\hline
 & 
\multicolumn{3}{c|}{\textbf{Used clang data}} & 
\multicolumn{1}{c|}{\textbf{bea-dev (full)}} \\ 

\textbf{\#} &  \textbf{Orig}   & \textbf{MF} & \textbf{ST}  & \textbf{F05 orig} \\ \hline

0  & no & no & no & 54.58\%  \\
1 & yes & no & no & 54.61\% ± 0.41\%  \\
2 & yes & no & yes & 54.52\% ± 0.48\%  \\
3  & yes & yes & no & 54.44\% ± 0.58\%  \\
4 & yes & yes & yes & 54.63\% ± 0.56\%  \\
\hline
\end{tabular}
}
\caption{\label{table-bea-dev-full} F0.5 on BEA-dev-full for GECToR fine-tunining experiments. For new experiments, average over all seeds ± 2 s.d. is shown.}
\end{table}

To evaluate the impact of adding the augmented data to the training dataset, we used an original subset of BEA dev and its augmented manually reviewed versions (described above). The results are shown in table \ref{table-bea-dev-556} and table \ref{table-bea-dev-195}.

\begin{table*}[h!]
\centering
\resizebox{0.75\textwidth}{!}{
\begin{tabular}{|c|ccc|ccc|}
\hline
 & 
\multicolumn{3}{c|}{\textbf{Used clang data}} & 
\multicolumn{3}{c|}{\textbf{bea-dev 556}} \\ 
\cline{2-7}

\textbf{\#} &  \textbf{Orig}   & \textbf{MF} & \textbf{ST}  & \textbf{F05 orig} & \textbf{F05 mf\_aug} & \textbf{Delta} \\ \hline
0 & no & no & no  & 59.23\%  & 58.96\% & -0.27\% \\
1 & yes & no & no &  57.79\% ± 0.82\% & 57.08\% ± 1.06\%  & -0.71\% \\
2 & yes & no & yes & 57.58\% ± 1.12\% & 57.01\% ± 1.2\%  & -0.57\% \\
3 & yes & yes & no & 57.88\% ± 0.7\% & 57.33\% ± 0.78\% & -0.55\% \\
4 & yes & yes & yes & 58.03\% ± 0.76\% & 57.5\% ± 1.04\% & -0.53\% \\

\hline

\end{tabular}
}
\caption{\label{table-bea-dev-556} F0.5 on BEA-dev-556 for GECToR fine-tunining experiments. For new experiments, average over all 10 seeds ± 2 s.d. is shown.}
\end{table*}

\begin{table*}[h]
\centering
\resizebox{0.75\textwidth}{!}{
\begin{tabular}{|c|ccc|ccc|}
\hline
 & 
\multicolumn{3}{c|}{\textbf{Used clang data}} & 
\multicolumn{3}{c|}{\textbf{bea-dev 195}} \\ 
\cline{2-7}

\textbf{\#} &  \textbf{Orig}   & \textbf{MF} & \textbf{ST}  & \textbf{F05 orig} & \textbf{F05 st\_aug} & \textbf{Delta} \\ \hline
0 & no & no & no & 58.28\%  & 48.74\% & -9.54\% \\
1 & yes & no & no & 56.33\% ± 2.1\% & 50.47\% ± 1.62\% & -5.86\%  \\
2 & yes & no & yes & 55.71\% ± 1.22\% & 54.31\% ± 1.62\% & -1.4\%  \\
3 & yes & yes & no & 55.77\% ± 1.04\% & 50.33\% ± 1.12\% & -5.44\% \\
4 & yes & yes & yes & 56.33\% ± 1.48\% & 54.86\% ± 1.46\%  & -1.47\%  \\
\hline

\end{tabular}
}
\caption{\label{table-bea-dev-195} F0.5 on BEA-dev-195 for GECToR fine-tunining experiments. For new experiments, average over all 10 seeds ± 2 s.d. is shown.}
\end{table*}

\subsubsection{Experiment with singular-they augmentation}
We can see that for the baseline model, the gap in F0.5 between the original and augmented (singular-they) version of BEA dev subset is quite significant -9.54\%. Fine-tuning on clang8 data led to the shrinking of the gap to -5.86\%. We think that this decrease illustrates not an improvement in gender bias, but rather a change in the baseline value due to a shift in precision/recall after fine-tuning. For a more fair comparison, we focused on analyzing the difference between the fine-tuned model on a combination of original and augmented data from clang8 (systems 2,3,4) and a model fine-tuned only on original clang data (system 1) (table \ref{table-bea-dev-195}).

We got an improvement in the F0.5 gap for systems 2 and 4 (from -5.86\% to -1.4\% and -1.47\% correspondingly). This reduction is driven by the improvement on the augmented version of bea-dev-195 subset (F0.5 +3.84\% and +4.39\%) without any (0\% for system 4) or with insignificant degradation in quality on the original bea-dev-195 subset (-0.62\% for system 3).

We also qualitatively examine the corrections to determine whether explicit instances of bias are reduced through data augmentation (table \ref{table-bea-dev-195-manual}). A linguist manually reviewed model predictions on bea-dev-195-st-aug for systems 1-4 and annotated predictions exhibiting explicit bias, which was defined as pluralization of a referent coreferring with singular \textit{they} or the replacement of singular \textit{they} with a gendered pronoun, or the replacement of \textit{themself} with \textit{themselves}. System 4 shows the greatest improvement with 7 cases over the baseline of 32. Examples of explicit bias are in Appendix \ref{sec:explicit}.

\begin{table}[h]
\centering
\resizebox{0.5\textwidth}{!}{
\begin{tabular}{|c|ccc|ccc|}
\hline
 & 
\multicolumn{3}{c|}{\textbf{Used clang data}} & 
\multicolumn{3}{c|}{\textbf{bea-dev-195-st-aug}} \\ 
\cline{2-7}

\textbf{\#} &  \textbf{Orig}   & \textbf{MF} & \textbf{ST}  & \textbf{\#} & \textbf{\# w/o refl} & \\ \hline

1 & yes & no & no & 32 & 29 & \\
2 & yes & no & yes & 8 & 7 & \\
3 & yes & yes & no & 34 & 30 & \\
4 & yes & yes & yes & 7 & 4 & \\
\hline

\end{tabular}
}
\caption{\label{table-bea-dev-195-manual} Number of sentences displaying explicit bias in bea-dev-195-st-aug. First column is total sentences found to have explicit bias, second is that count minus cases of "themself">"themselves".}
\end{table}

\subsubsection{Experiment with feminine/masculine augmentation}

For feminine/masculine augmentation, the initial gap between the original subset of BEA (556 sentences) and the augmented version is much smaller -0.71\%. Fine-tuning on original and feminine/masculine augmentation data (system 3) very slightly reduces this difference only to -0.55\%. It’s interesting that even singular-they augmentation, without any other gender-swapping, seems to provide a very similar result (difference of -0.57\%). However, given the size of confidence intervals, we cannot say that any of our experiments had a significant impact on the gap.

\subsubsection{Experiment with both augmentations}
Finally, we tried to apply both kinds of augmentation - singular-they and feminine/masculine CDA. The resulting model (system 4) is producing very similar results in terms of gap difference for both BEA subsets that we used: -1.47\% on bea-195 (system 2 gap is -1.4\%) and -0.53\% on bea-556 (system 3 gap is -0.55\%), which is showing that multiple biases might be handled with such a single fine-tuning approach at once.
It also seems that augmented training data of two kinds does not interfere with any one evaluation but also does not provide additional benefits from this data interaction.

We believe that there are many other potential possibilities to incorporate augmented data into different stages of the training or change the proportion or the absolute number of original and augmented sentences in training data that might lead to even better improvement with little to no quality degradation. We would like to explore some of them in future work. 

\section{Conclusion}

In this work, we developed a novel technique for data augmentation with sentences containing \textit{they} that has an unambiguous singular reference and applied it to the GEC case. We used this technique to help develop a dataset of singular \textit{they} data to parallel data in the BEA shared task dataset that has masculine and feminine pronouns, and with this, we show that GEC systems display bias in their treatment of singular \textit{they} sentences compared to sentences with masculine or feminine pronouns. Additionally, we demonstrated that this technique could be used to reduce bias in GEC systems by fine-tuning the GEC system on the generated synthetic training data.

Because this technique is data-oriented, we believe that it has wider applications, and other NLP systems that display degraded performance with respect to singular \textit{they} may benefit from being trained on data created through this technique.

\section*{Limitations}
As noted, this work is limited in that it does not address neopronouns. We speculate that the augmentation techniques deployed in this work may extend to these pronouns as well, we recognize that they do not have the same linguistic reality as \textit{he}/\textit{she}/\textit{they} pronouns. Neopronouns may be similar to singular \textit{they} in being relatively infrequent in a naturalistic corpus, but they are also different in that they don't overlap with a frequent morphologically-identical paradigm like plural \textit{they}.

Additionally, the singular \textit{they} augmentation technique we propose is specific to English and distributional facts about English pronouns. For one, English singular \textit{they} morphologically overlaps with a plural pronoun, which is the primary motivation for using coreference information to identify contexts where \textit{they} would have a primarily singular interpretation. This is often not the case for other languages, as in Swedish where the gender-neutral \textit{hen} is functionally similar to singular \textit{they} but morphologically and distributionally dissimilar in that it does not overlap with a plural pronoun \citep{gustafsson_senden_introducing_2015}. 

\section*{Ethics Statement}

\subsection*{Dataset risks}

We do not anticipate any risks in releasing the evaluation dataset. This dataset was constructed through the modification of a publicly available dataset commonly used in the evaluation of GEC systems, the dev set of the BEA-2019 shared task \citep{bryant_bea-2019_2019}. These modifications involve the change of gendered words and agreeing verbs to create parallel data across masculine, feminine, and singular \textit{they} pronouns with the goal of evaluating bias in GEC systems. By enabling researchers to measure bias in this way, we believe that the release of this dataset will aid further study in reducing bias in these systems by providing a benchmark.

\subsection*{Risks of describing data augmentation techniques}

We caution that the singular \textit{they} data augmentation technique used in this paper was not designed to generate text that surfaces directly to users. There may be risks to deploying data augmentation techniques at runtime as these techniques are designed to modify gender identity terms; depending on the context of deployment, users may be harmed by such modifications if they result in misgendering or erasure. On the other hand, as we show in this work, use of these techniques to generate training data can reduce bias, and we believe that in this way, the description of this technique will aid in reducing bias in NLP systems.

\section*{Acknowledgements}

This research was supported by Grammarly. We thank our colleagues Leonardo Neves, Jade Razzaghi, Yichen Mo, Serhii Yavnyi, and Knar Hovakimyan for their brilliant insights and suggestions over the course of this work. We would also like to thank 3 anonymous reviewers for their helpful comments.

\bibliography{library, paper}

\begin{thebibliography}{32}
\expandafter\ifx\csname natexlab\endcsname\relax\def\natexlab#1{#1}\fi

\bibitem[{Ackerman(2019)}]{ackerman_syntactic_2019}
Lauren Ackerman. 2019.
\newblock \href {https://doi.org/10.5334/gjgl.721} {Syntactic and cognitive
  issues in investigating gendered coreference}.
\newblock \emph{Glossa: a journal of general linguistics}, 4(1).
\newblock Number: 1 Publisher: Open Library of Humanities.

\bibitem[{Baumler and Rudinger(2022)}]{baumler_recognition_2022}
Connor Baumler and Rachel Rudinger. 2022.
\newblock \href {https://aclanthology.org/2022.naacl-main.250} {Recognition of
  {They}/{Them} as {Singular} {Personal} {Pronouns} in {Coreference}
  {Resolution}}.
\newblock In \emph{Proceedings of the 2022 {Conference} of the {North}
  {American} {Chapter} of the {Association} for {Computational} {Linguistics}:
  {Human} {Language} {Technologies}}, pages 3426--3432, Seattle, United States.
  Association for Computational Linguistics.

\bibitem[{Bjorkman(2017)}]{bjorkman_singular_2017}
Bronwyn~M. Bjorkman. 2017.
\newblock \href {https://doi.org/10.5334/gjgl.374} {Singular they and the
  syntactic representation of gender in {English}}.
\newblock \emph{Glossa: a journal of general linguistics}, 2(1).
\newblock Number: 1 Publisher: Open Library of Humanities.

\bibitem[{Blodgett et~al.(2020)Blodgett, Barocas, Daumé~III, and
  Wallach}]{blodgett_language_2020}
Su~Lin Blodgett, Solon Barocas, Hal Daumé~III, and Hanna Wallach. 2020.
\newblock \href {https://arxiv.org/abs/2005.14050v2} {Language ({Technology})
  is {Power}: {A} {Critical} {Survey} of "{Bias}" in {NLP}}.

\bibitem[{Bolukbasi et~al.(2016)Bolukbasi, Chang, Zou, Saligrama, and
  Kalai}]{bolukbasi_man_2016}
Tolga Bolukbasi, Kai-Wei Chang, James Zou, Venkatesh Saligrama, and Adam Kalai.
  2016.
\newblock \href {https://doi.org/10.48550/arXiv.1607.06520} {Man is to
  {Computer} {Programmer} as {Woman} is to {Homemaker}? {Debiasing} {Word}
  {Embeddings}}.
\newblock Number: arXiv:1607.06520 arXiv:1607.06520 [cs, stat].

\bibitem[{Bryant et~al.(2019)Bryant, Felice, Andersen, and
  Briscoe}]{bryant_bea-2019_2019}
Christopher Bryant, Mariano Felice, Øistein~E. Andersen, and Ted Briscoe.
  2019.
\newblock \href {https://doi.org/10.18653/v1/W19-4406} {The {BEA}-2019 {Shared}
  {Task} on {Grammatical} {Error} {Correction}}.
\newblock In \emph{Proceedings of the {Fourteenth} {Workshop} on {Innovative}
  {Use} of {NLP} for {Building} {Educational} {Applications}}, pages 52--75,
  Florence, Italy. Association for Computational Linguistics.

\bibitem[{Bryant et~al.(2022)Bryant, Yuan, Qorib, Cao, Ng, and
  Briscoe}]{bryant_grammatical_2022}
Christopher Bryant, Zheng Yuan, Muhammad~Reza Qorib, Hannan Cao, Hwee~Tou Ng,
  and Ted Briscoe. 2022.
\newblock \href {https://doi.org/10.48550/arXiv.2211.05166} {Grammatical
  {Error} {Correction}: {A} {Survey} of the {State} of the {Art}}.
\newblock ArXiv:2211.05166 [cs].

\bibitem[{Cao and Daumé~III(2021)}]{cao_toward_2021}
Yang~Trista Cao and Hal Daumé~III. 2021.
\newblock \href {https://doi.org/10.1162/coli_a_00413} {Toward
  {Gender}-{Inclusive} {Coreference} {Resolution}: {An} {Analysis} of {Gender}
  and {Bias} {Throughout} the {Machine} {Learning} {Lifecycle}*}.
\newblock \emph{Computational Linguistics}, 47(3):615--661.
\newblock Place: Cambridge, MA Publisher: MIT Press.

\bibitem[{Dev et~al.(2021)Dev, Monajatipoor, Ovalle, Subramonian, Phillips, and
  Chang}]{dev_harms_2021}
Sunipa Dev, Masoud Monajatipoor, Anaelia Ovalle, Arjun Subramonian, J.~M.
  Phillips, and Kai~Wei Chang. 2021.
\newblock \href {https://doi.org/10.18653/v1/2021.emnlp-main.150} {Harms of
  {Gender} {Exclusivity} and {Challenges} in {Non}-{Binary} {Representation} in
  {Language} {Technologies}}.
\newblock In \emph{{EMNLP}}.

\bibitem[{Devinney et~al.(2022)Devinney, Björklund, and
  Björklund}]{devinney_theories_2022}
Hannah Devinney, Jenny Björklund, and Henrik Björklund. 2022.
\newblock \href {https://doi.org/10.1145/3531146.3534627} {Theories of
  “{Gender}” in {NLP} {Bias} {Research}}.
\newblock In \emph{2022 {ACM} {Conference} on {Fairness}, {Accountability}, and
  {Transparency}}, {FAccT} '22, pages 2083--2102, New York, NY, USA.
  Association for Computing Machinery.

\bibitem[{Gustafsson~Sendén et~al.(2015)Gustafsson~Sendén, Bäck, and
  Lindqvist}]{gustafsson_senden_introducing_2015}
Marie Gustafsson~Sendén, Emma~A. Bäck, and Anna Lindqvist. 2015.
\newblock \href {https://www.frontiersin.org/articles/10.3389/fpsyg.2015.00893}
  {Introducing a gender-neutral pronoun in a natural gender language: the
  influence of time on attitudes and behavior}.
\newblock \emph{Frontiers in Psychology}, 6.

\bibitem[{Han and Moulton(2022)}]{han_processing_2022}
Chung-hye Han and Keir Moulton. 2022.
\newblock \href {https://doi.org/10.1017/cnj.2022.30} {Processing
  bound-variable singular they}.
\newblock \emph{Canadian Journal of Linguistics/Revue canadienne de
  linguistique}, 67(3):267--301.
\newblock Publisher: Cambridge University Press.

\bibitem[{Konnelly and Cowper(2020)}]{konnelly_gender_2020}
Lex Konnelly and Elizabeth Cowper. 2020.
\newblock \href {https://doi.org/10.5334/gjgl.1000} {Gender diversity and
  morphosyntax: {An} account of singular they}.
\newblock \emph{Glossa: a journal of general linguistics}, 5(1).
\newblock Number: 1 Publisher: Open Library of Humanities.

\bibitem[{Levesque et~al.(2012)Levesque, Davis, and
  Morgenstern}]{levesque_winograd_2012}
Hector~J. Levesque, Ernest Davis, and Leora Morgenstern. 2012.
\newblock The {Winograd} schema challenge.
\newblock In \emph{Proceedings of the {Thirteenth} {International} {Conference}
  on {Principles} of {Knowledge} {Representation} and {Reasoning}}, {KR}'12,
  pages 552--561, Rome, Italy. AAAI Press.

\bibitem[{Lewis et~al.(2020)Lewis, Liu, Goyal, Ghazvininejad, Mohamed, Levy,
  Stoyanov, and Zettlemoyer}]{lewis-etal-2020-bart}
Mike Lewis, Yinhan Liu, Naman Goyal, Marjan Ghazvininejad, Abdelrahman Mohamed,
  Omer Levy, Veselin Stoyanov, and Luke Zettlemoyer. 2020.
\newblock \href {https://doi.org/10.18653/v1/2020.acl-main.703} {{BART}:
  Denoising sequence-to-sequence pre-training for natural language generation,
  translation, and comprehension}.
\newblock In \emph{Proceedings of the 58th Annual Meeting of the Association
  for Computational Linguistics}, pages 7871--7880, Online. Association for
  Computational Linguistics.

\bibitem[{Liu et~al.(2019)Liu, Ott, Goyal, Du, Joshi, Chen, Levy, Lewis,
  Zettlemoyer, and Stoyanov}]{liu_roberta_2019}
Yinhan Liu, Myle Ott, Naman Goyal, Jingfei Du, Mandar Joshi, Danqi Chen, Omer
  Levy, Mike Lewis, Luke Zettlemoyer, and Veselin Stoyanov. 2019.
\newblock \href {https://doi.org/10.48550/arXiv.1907.11692} {{RoBERTa}: {A}
  {Robustly} {Optimized} {BERT} {Pretraining} {Approach}}.
\newblock ArXiv:1907.11692 [cs].

\bibitem[{Lu et~al.(2019)Lu, Mardziel, Wu, Amancharla, and
  Datta}]{lu_gender_2019}
Kaiji Lu, Piotr Mardziel, Fangjing Wu, Preetam Amancharla, and Anupam Datta.
  2019.
\newblock \href {http://arxiv.org/abs/1807.11714} {Gender {Bias} in {Neural}
  {Natural} {Language} {Processing}}.
\newblock \emph{arXiv:1807.11714 [cs]}.
\newblock ArXiv: 1807.11714.

\bibitem[{Maudslay et~al.(2020)Maudslay, Gonen, Cotterell, and
  Teufel}]{maudslay_its_2020}
Rowan~Hall Maudslay, Hila Gonen, Ryan Cotterell, and Simone Teufel. 2020.
\newblock \href {http://arxiv.org/abs/1909.00871} {It's {All} in the {Name}:
  {Mitigating} {Gender} {Bias} with {Name}-{Based} {Counterfactual} {Data}
  {Substitution}}.
\newblock \emph{arXiv:1909.00871 [cs]}.
\newblock ArXiv: 1909.00871.

\bibitem[{McConnell-Ginet(2013)}]{mcconnell-ginet__2013}
Sally McConnell-Ginet. 2013.
\newblock \href {https://doi.org/10.1515/9783110307337.3} {` {Gender} and its
  relation to sex: {The} myth of ‘natural’ gender}.
\newblock In \emph{` {Gender} and its relation to sex: {The} myth of
  ‘natural’ gender}, pages 3--38. De Gruyter Mouton.

\bibitem[{Mizumoto et~al.(2011)Mizumoto, Komachi, Nagata, and
  Matsumoto}]{mizumoto_mining_2011}
Tomoya Mizumoto, Mamoru Komachi, Masaaki Nagata, and Yuji Matsumoto. 2011.
\newblock \href {https://aclanthology.org/I11-1017} {Mining {Revision} {Log} of
  {Language} {Learning} {SNS} for {Automated} {Japanese} {Error} {Correction}
  of {Second} {Language} {Learners}}.
\newblock In \emph{Proceedings of 5th {International} {Joint} {Conference} on
  {Natural} {Language} {Processing}}, pages 147--155, Chiang Mai, Thailand.
  Asian Federation of Natural Language Processing.

\bibitem[{Moulton et~al.(2020)Moulton, Han, Block, Gendron, and
  Nederveen}]{moulton_singular_2020}
Keir Moulton, Chung-hye Han, Trevor Block, Holly Gendron, and Sander Nederveen.
  2020.
\newblock \href {https://doi.org/10.5334/gjgl.1012} {Singular they in context}.
\newblock \emph{Glossa: a journal of general linguistics}, 5(1).
\newblock Number: 1 Publisher: Open Library of Humanities.

\bibitem[{Omelianchuk et~al.(2020)Omelianchuk, Atrasevych, Chernodub, and
  Skurzhanskyi}]{omelianchuk_gector_2020}
Kostiantyn Omelianchuk, Vitaliy Atrasevych, Artem Chernodub, and Oleksandr
  Skurzhanskyi. 2020.
\newblock \href {http://arxiv.org/abs/2005.12592} {{GECToR} -- {Grammatical}
  {Error} {Correction}: {Tag}, {Not} {Rewrite}}.
\newblock ArXiv:2005.12592 [cs].

\bibitem[{Omelianchuk et~al.(2021)Omelianchuk, Raheja, and
  Skurzhanskyi}]{omelianchuk_text_2021}
Kostiantyn Omelianchuk, Vipul Raheja, and Oleksandr Skurzhanskyi. 2021.
\newblock \href {https://aclanthology.org/2021.bea-1.2} {Text {Simplification}
  by {Tagging}}.
\newblock In \emph{Proceedings of the 16th {Workshop} on {Innovative} {Use} of
  {NLP} for {Building} {Educational} {Applications}}, pages 11--25, Online.
  Association for Computational Linguistics.

\bibitem[{Rothe et~al.(2021)Rothe, Mallinson, Malmi, Krause, and
  Severyn}]{rothe_simple_2021}
Sascha Rothe, Jonathan Mallinson, Eric Malmi, Sebastian Krause, and Aliaksei
  Severyn. 2021.
\newblock \href {https://doi.org/10.18653/v1/2021.acl-short.89} {A {Simple}
  {Recipe} for {Multilingual} {Grammatical} {Error} {Correction}}.
\newblock In \emph{Proceedings of the 59th {Annual} {Meeting} of the
  {Association} for {Computational} {Linguistics} and the 11th {International}
  {Joint} {Conference} on {Natural} {Language} {Processing} ({Volume} 2:
  {Short} {Papers})}, pages 702--707, Online. Association for Computational
  Linguistics.

\bibitem[{Rudinger et~al.(2018)Rudinger, Naradowsky, Leonard, and
  Van~Durme}]{rudinger_gender_2018}
Rachel Rudinger, Jason Naradowsky, Brian Leonard, and Benjamin Van~Durme. 2018.
\newblock \href {http://arxiv.org/abs/1804.09301} {Gender {Bias} in
  {Coreference} {Resolution}}.
\newblock \emph{arXiv:1804.09301 [cs]}.
\newblock ArXiv: 1804.09301.

\bibitem[{Saunders and Byrne(2020)}]{saunders-byrne-2020-reducing}
Danielle Saunders and Bill Byrne. 2020.
\newblock \href {https://doi.org/10.18653/v1/2020.acl-main.690} {Reducing
  gender bias in neural machine translation as a domain adaptation problem}.
\newblock In \emph{Proceedings of the 58th Annual Meeting of the Association
  for Computational Linguistics}, pages 7724--7736, Online. Association for
  Computational Linguistics.

\bibitem[{Sorokin(2022)}]{sorokin-2022-improved}
Alexey Sorokin. 2022.
\newblock \href {https://aclanthology.org/2022.emnlp-main.785} {Improved
  grammatical error correction by ranking elementary edits}.
\newblock In \emph{Proceedings of the 2022 Conference on Empirical Methods in
  Natural Language Processing}, pages 11416--11429, Abu Dhabi, United Arab
  Emirates. Association for Computational Linguistics.

\bibitem[{Strunk and White(1999)}]{strunk_elements_1999}
William Strunk and E.~B. White. 1999.
\newblock \emph{The elements of style}, 4th ed edition.
\newblock Allyn and Bacon, Boston.

\bibitem[{Sun et~al.(2021)Sun, Webster, Shah, Wang, and
  Johnson}]{sun_they_2021}
Tony Sun, Kellie Webster, Apu Shah, William~Yang Wang, and Melvin Johnson.
  2021.
\newblock \href {https://doi.org/10.48550/arXiv.2102.06788} {They, {Them},
  {Theirs}: {Rewriting} with {Gender}-{Neutral} {English}}.
\newblock ArXiv:2102.06788 [cs].

\bibitem[{Vanmassenhove et~al.(2021)Vanmassenhove, Emmery, and
  Shterionov}]{vanmassenhove_neutral_2021}
Eva Vanmassenhove, Chris Emmery, and Dimitar Shterionov. 2021.
\newblock \href {https://doi.org/10.18653/v1/2021.emnlp-main.704} {{NeuTral}
  {Rewriter}: {A} {Rule}-{Based} and {Neural} {Approach} to {Automatic}
  {Rewriting} into {Gender} {Neutral} {Alternatives}}.
\newblock In \emph{Proceedings of the 2021 {Conference} on {Empirical}
  {Methods} in {Natural} {Language} {Processing}}, pages 8940--8948, Online and
  Punta Cana, Dominican Republic. Association for Computational Linguistics.

\bibitem[{Wang et~al.(2022)Wang, Rubinstein, and Cohn}]{wang_measuring_2022}
Jun Wang, Benjamin Rubinstein, and Trevor Cohn. 2022.
\newblock \href {https://doi.org/10.18653/v1/2022.acl-long.184} {Measuring and
  {Mitigating} {Name} {Biases} in {Neural} {Machine} {Translation}}.
\newblock In \emph{Proceedings of the 60th {Annual} {Meeting} of the
  {Association} for {Computational} {Linguistics} ({Volume} 1: {Long}
  {Papers})}, pages 2576--2590, Dublin, Ireland. Association for Computational
  Linguistics.

\bibitem[{Zhao et~al.(2018)Zhao, Wang, Yatskar, Ordonez, and
  Chang}]{zhao_gender_2018}
Jieyu Zhao, Tianlu Wang, Mark Yatskar, Vicente Ordonez, and Kai-Wei Chang.
  2018.
\newblock \href {http://arxiv.org/abs/1804.06876} {Gender {Bias} in
  {Coreference} {Resolution}: {Evaluation} and {Debiasing} {Methods}}.
\newblock \emph{arXiv:1804.06876 [cs]}.
\newblock ArXiv: 1804.06876.

\end{thebibliography}
\bibliographystyle{acl_natbib}

\appendix

\section{Examples of explicit bias}\label{sec:explicit}

Here we present examples of how the systems investigated here exhibit explicit bias. These are indicated in \textbf{bold text}. The list numbers correspond to the systems indicated in Table \ref{table-bea-dev-full}.\\

\noindent Source text:\\ \texttt{I said that because when I was eleven my best friend had been operated on \textbf{their} back and before the operation he came with me and every day I had to wait for them because they spendt a lot of time in the shower cleaning their long hais , I hated that !}

\begin{enumerate}
\setcounter{enumi}{-1}
    \item \texttt{I said that because when I was eleven my best friend had been operated on and before the operation he came with me and every day I had to wait for them because they spendt a lot of time in the shower cleaning their long hair . I hated that !}
    \item \texttt{I said that because when I was eleven my best friend had been operated on \textbf{his} back and before the operation he came with me and every day I had to wait for them because they spendt a lot of time in the shower cleaning their long hair . I hated that !}
    \item \texttt{I said that because when I was eleven my best friend had been operated on their back and before the operation he came with me and every day I had to wait for them because they spendt a lot of time in the shower cleaning their long hair . I hated that !}
    \item \texttt{I said that because when I was eleven my best friend had been operated on their back and before the operation he came with me and every day I had to wait for them because they spendt a lot of time in the shower cleaning their long hair . I hated that !}
    \item \texttt{I said that because when I was eleven my best friend had been operated on their back and before the operation he came with me and every day I had to wait for them because they spendt a lot of time in the shower cleaning their long hair . I hated that !}
\end{enumerate}

\noindent Source text:\\ \texttt{To take my friend , Taylor , as an example , their homeland were over - developed and become an arid place when they were \textbf{a child} .}

\begin{enumerate}
\setcounter{enumi}{-1}
    \item \texttt{To take my friend , Taylor , as an example , their homeland was over - developed and became an arid place when they were \textbf{children} .}
    \item \texttt{To take my friend , Taylor , as an example , their homeland was over - developed and became an arid place when they were \textbf{children} .}
    \item \texttt{To take my friend , Taylor , as an example , their homeland was over - developed and became an arid place when they were a child .}
    \item \texttt{To take my friend , Taylor , as an example , their homeland was over - developed and became an arid place when they were \textbf{children} .}
    \item \texttt{To take my friend , Taylor , as an example , their homeland was over - developed and became an arid place when they were a child .}
\end{enumerate}

\noindent Source text:\\ \texttt{My father has the same program in \textbf{their} computer and is able to buy on line tickets w / out leaving the house and moreover w / out picking up the phone .}

\begin{enumerate}
\setcounter{enumi}{-1}
    \item \texttt{My father has the same program on \textbf{his} computer and is able to buy online tickets w / out leaving the house and moreover w / out picking up the phone .}
    \item \texttt{My father has the same program on \textbf{his} computer and is able to buy online tickets w / out leaving the house and moreover w / out picking up the phone .}
    \item \texttt{My father has the same program on their computer and is able to buy online tickets without leaving the house and moreover without picking up the phone .}
    \item \texttt{My father has the same program on \textbf{his} computer and is able to buy online tickets w / out leaving the house and moreover w / out picking up the phone .}
    \item \texttt{My father has the same program on their computer and is able to buy online tickets without leaving the house and moreover without picking up the phone .}
\end{enumerate}

\clearpage
\onecolumn
\section{Error distribution on BEA-dev subsets}
\subsection{Edit type distribution on "195" and "556" subsets of bea\_dev}\label{sec:errortypedist}

\begin{tabular}{|c|cc|cc|cc|cc|}
\hline
 &  \multicolumn{4}{|c|}{\textbf{bea\_dev\_195}} & \multicolumn{4}{|c|}{\textbf{bea\_dev\_556}} \\
 &  \multicolumn{2}{|c|}{\textbf{orig}} & \multicolumn{2}{c|}{\textbf{st\_aug}} & \multicolumn{2}{c|}{\textbf{orig}} & \multicolumn{2}{c|}{\textbf{mf\_aug}} \\
 \hline
Edit type & \# edits & \% edits & \# edits & \% edits & \# edits & \% edits & \# edits & \% edits \\
R (replacement) & 374 & 66.1\% & 371 & 66.0\% & 814 & 64.1\% &  814 & 64.1\% \\
M (missing) & 156 & 27.6\% & 157 & 27.9\% & 358 & 28.2\% & 358 & 28.2\%  \\
U (unnecessary) & 36 & 6.4\% & 34 & 6.0\% & 97 & 7.6\% & 96 & 7.6\% \\
    \hline
Total & 566 & 100\% & 562 & 100\% & 1269 & 100\% & 1268 & 100\% \\
\hline
\end{tabular}

\vspace{0.25in}
\subsection{Error categories distribution on "195" and "556" subsets of bea\_dev}\label{sec:errorcategory}

\begin{tabular}{|c|cc|cc|cc|cc|}
\hline
 &  \multicolumn{4}{|c|}{bea\_dev\_195} & \multicolumn{4}{|c|}{bea\_dev\_556} \\
 &  \multicolumn{2}{|c|}{orig} & \multicolumn{2}{c|}{st\_aug} & \multicolumn{2}{c|}{orig} & \multicolumn{2}{c|}{mf\_aug} \\
 \hline
Error category & \# edits & \% edits & \# edits & \% edits & \# edits & \% edits & \# edits & \% edits \\

PUNCT & 150 & 26.5\% & 150 & 26.7\%  & 314 & 24.7\% & 314 & 24.8\%  \\
    VERB:TENSE & 68 & 12.0\% & 66 & 11.7\%  & 139 & 11.0\% & 139 & 11.0\%  \\
    OTHER & 45 & 8.0\% & 48 & 8.5\%  & 121 & 9.5\% & 122 & 9.6\%  \\
    PREP & 46 & 8.1\% & 46 & 8.2\%  & 115 & 9.1\% & 114 & 9.0\%  \\
    DET & 35 & 6.2\% & 34 & 6.0\%  & 92 & 7.2\% & 89 & 7.0\%  \\
    ORTH & 44 & 7.8\% & 44 & 7.8\%  & 84 & 6.6\% & 84 & 6.6\%  \\
    SPELL & 42 & 7.4\% & 41 & 7.3\%  & 73 & 5.8\% & 74 & 5.8\%  \\
    VERB & 31 & 5.5\% & 30 & 5.3\%  & 67 & 5.3\% & 67 & 5.3\%  \\
    VERB:FORM & 14 & 2.5\% & 15 & 2.7\%  & 36 & 2.8\% & 37 & 2.9\%  \\
    PRON & 10 & 1.8\% & 10 & 1.8\%  & 35 & 2.8\% & 36 & 2.8\%  \\
    NOUN & 17 & 3.0\% & 17 & 3.0\%  & 34 & 2.7\% & 35 & 2.8\%  \\
    NOUN:NUM & 6 & 1.1\% & 6 & 1.1\%  & 33 & 2.6\% & 32 & 2.5\%  \\
    VERB:SVA & 12 & 2.1\% & 11 & 2.0\%  & 25 & 2.0\% & 25 & 2.0\%  \\
    MORPH & 8 & 1.4\% & 7 & 1.2\%  & 24 & 1.9\% & 23 & 1.8\%  \\
    ADV & 12 & 2.1\% & 12 & 2.1\%  & 16 & 1.3\% & 17 & 1.3\%  \\
    ADJ & 10 & 1.8\% & 9 & 1.6\%  & 17 & 1.3\% & 16 & 1.3\%  \\
    WO & 5 & 0.9\% & 5 & 0.9\%  & 12 & 0.9\% & 12 & 0.9\%  \\
    NOUN:POSS & 3 & 0.5\% & 3 & 0.5\%  & 10 & 0.8\% & 10 & 0.8\%  \\
    PART & 5 & 0.9\% & 5 & 0.9\%  & 8 & 0.6\% & 8 & 0.6\%  \\
    CONTR & 1 & 0.2\% & 1 & 0.2\%  & 6 & 0.5\% & 6 & 0.5\%  \\
    CONJ & 2 & 0.4\% & 2 & 0.4\%  & 6 & 0.5\% & 6 & 0.5\%  \\
    VERB:INFL & 0 & 0.0\% & 0 & 0.0\%  & 2 & 0.2\% & 2 & 0.2\%  \\
    \hline
    Total & 566 & 100\% & 562 & 100\% & 1269 & 100\% & 1268 & 100\% \\

\hline
\end{tabular}

\clearpage
\section{Hyperparameter values for the fine-tuning of GECToR}

\begin{tabular}{l|l}
    \hline
    \textbf{Hyperparameter name} & \textbf{Hyperparameter value} \\
    \hline
    batch\_size & 32 \\
    accumulation\_size & 4 \\
    n\_epoch & 5 \\
    patience & 3 \\
    max\_len & 5 \\
    lr & 1e-05 \\
    cold\_steps\_count & 1 \\
    cold\_lr & 0.001 \\
    tp\_prob & 1 \\
    tn\_prob & 1 \\
    updates\_per\_epoch & 10000 \\
    special\_tokens\_fix & 1 \\
    transformer\_model & roberta-base \\
    Pretrained model & \\
    Inference tweaks: & \\ 
    minimum error probability & 0.5 \\
    Inference tweaks: & \\ 
    confidence & 0.2 \\
    \hline
\end{tabular}

\section{Hyperparameter values for the fine-tuning of BART}

\begin{tabular}{l|l}
    \hline
    \textbf{Hyperparameter name} & \textbf{Hyperparameter value} \\
    \hline
    base\_model & BART-Large \\
    src\_max\_length & 80 \\
    tgt\_max\_length & 85 \\
    beam & 2 \\
    max\_update & 16000 \\
    loss\_criterion & label\_smoothed\_cross\_entropy \\
    optimizer & Adam \\
    weight\_decay & 0.0 \\
    adam\_betas & (0.9, 0.98) \\
    adam\_eps & 1e-06 \\
    lr & 3e-05 \\
    \hline
\end{tabular}

\clearpage
\section{Full results of evaluation}

\begin{table*}[h]
\centering
\resizebox{\textwidth}{!}{
\begin{tabular}{|c|ccc|ccc|ccc|ccc|ccc|}
\hline
System & 
\multicolumn{3}{c|}{\textbf{bea-dev-full}} & 
\multicolumn{3}{c|}{\textbf{bea-195-orig}} & \multicolumn{3}{c|}{\textbf{bea-195-st-aug}} &
\multicolumn{3}{c|}{\textbf{bea-556-orig}} & \multicolumn{3}{c|}{\textbf{bea--556-mf-aug}}
\\ 
\cline{2-16}

 &  
\textbf{P}   & \textbf{R} & \textbf{F05}  & \textbf{P}   & \textbf{R} & \textbf{F05} & 
\textbf{P}   & \textbf{R} & \textbf{F05} &
\textbf{P}   & \textbf{R} & \textbf{F05} &
\textbf{P}   & \textbf{R} & \textbf{F05} \\ \hline

GECToR (roberta-base) & 64.05\% & 34.28\% & 54.57\% & 70.11\% & 34.81\% & 58.28\% & 56.33\%  & 31.67\%  & 48.74\% & 69.46\%  & 37.27\%  & 59.23\% & 68.84\% & 37.46\%  & 58.96\% \\

BART (seq2seq)  & 57.46\%  & 39.7\% & 52.74\% & 61.32\% & 42.58\% & 56.63\% & 53.59\%  & 39.86\%  & 50.13\%  &  62.73\% & 46.41\% & 58.61\%   & 63.05\%  & 46.29\% & 58.79\%  \\

EditScorer (roberta-large)  & 70.29\% & 35.77\% & 58.92\% & 73.98\% & 35.16\% & 60.6\% & 63.76\% & 33.81\% & 54.16\% & 75\% & 37.59\% & 62.55\%  & 73.42\%  & 37.46\% & 61.59\% \\

\hline

\end{tabular}
}
\caption{\label{full-table-sota-overview}
Scores on BEA-dev subsets for strong GEC baselines.}
\end{table*}

\begin{table*}[h]
\centering
\resizebox{\textwidth}{!}{
\begin{tabular}{|c|ccc|ccc|}
\hline
 & 
\multicolumn{3}{c|}{\textbf{Used clang data}} & 
\multicolumn{3}{c|}{\textbf{bea-dev (full)}} \\ 

\textbf{\#} &  \textbf{Orig}   & \textbf{MF} & \textbf{ST}  & \textbf{Precision} & \textbf{Recall} & \textbf{F05 orig} \\ \hline

0  & no & no & no & 64.05\% & 34.28\% & 54.58\%  \\
1 & yes & no & no & 62.29\% ± 1.3\% & 36.6\% ± 1.58\% & 54.61\% ± 0.41\%  \\
2 & yes & no & yes & 62.19\% ± 1.12\% & 36.35\% ± 0.98\% & 54.52\% ± 0.48\%  \\
3  & yes & yes & no & 62.38\% ± 1.1\% & 36.25\% ± 1.34\% & 54.44\% ± 0.58\%  \\
4 & yes & yes & yes & 62.41\% ± 0.72\% & 36.46\% ± 0.82\% & 54.63\% ± 0.56\%  \\

\hline
\end{tabular}
}
\caption{\label{table-bea-dev-full-precision-recall} F0.5 on BEA-dev-full for GECToR fine-tunining experiments. For new experiments, average over all seeds ± 2 s.d. is shown.}
\end{table*}

\end{document}